\documentclass[runningheads]{llncs}

\usepackage{eccv}

\usepackage{wrapfig}
\usepackage{eccvabbrv}
\usepackage{multirow}
\usepackage{makecell}
\usepackage[table]{xcolor}
\usepackage{pifont}
\newcommand{\cmark}{\ding{51}}
\newcommand{\xmark}{\ding{55}}
\usepackage{graphicx}
\usepackage{booktabs}

\usepackage[accsupp]{axessibility}  

\usepackage{hyperref}
\usepackage{orcidlink}
\usepackage{mdframed}
\usepackage{listings}
\usepackage{xcolor}
\usepackage{placeins}
\usepackage{tabularx}
\usepackage{array}
\usepackage[most]{tcolorbox}
\usepackage{xcolor}
\usepackage{booktabs}

\definecolor{uiheader}{HTML}{2D3748}
\definecolor{uibg}{HTML}{F7FAFC}
\definecolor{uitext}{HTML}{4A5568}
\definecolor{uiblue}{HTML}{3182CE}
\definecolor{uired}{HTML}{E53E3E}
\definecolor{uigreen}{HTML}{38A169}

\newcommand{\genetag}[2][gray!20]{%
    \tcbox[on line, boxsep=0pt, left=3pt, right=3pt, top=2pt, bottom=2pt, 
           colback=#1, colframe=#1, arc=2pt, fontupper=\sffamily\scriptsize\bfseries, 
           colupper=black!80]{#2}%
}
\usepackage{float}
\usepackage[capitalize]{cleveref}
\begin{document}

\title{CLARITY: Medical World Model for Guiding Treatment Decisions by Simulating Context-Aware Disease Trajectories}
\author{
Tianxingjian Ding \and
Yuanhao Zou \and
Chen Chen \and
Mubarak Shah \and
Yu Tian
}
\authorrunning{Ding et al.}
\titlerunning{CLARITY: Medical World Model for Treatment Decisions}
\institute{
Institute of Artificial Intelligence, University of Central Florida \\
\url{https://dingtianxingjian.github.io/clarity-project-page/}
}
\maketitle

\begin{abstract}

Clinical decision-making in oncology requires forecasting how disease evolves under treatment, yet most AI systems remain static predictors that cannot model longitudinal, treatment-conditioned progression. Although generative and world models have demonstrated strong capabilities in general domains, their adaptation to medicine remains limited and insufficient for capturing complex, treatment-induced physiological dynamics across temporal scales.
To address these gaps, we introduce CLARITY, a medical world model that enables counterfactual simulation of treatment-conditioned disease trajectories for clinical decision-making. By jointly encoding imaging-derived latent states, temporal intervals that capture irregular follow-ups, and patient-specific clinical context, CLARITY learns smooth and interpretable representations of disease progression, allowing the model to simulate how alternative treatments reshape future disease dynamics. Because treatment optimization is inherently sequential and uncertain, requiring evaluation of long-term outcomes across multiple possible interventions, we further propose an entropy-regularized, computationally efficient long-horizon prediction-to-decision framework that plans treatment strategies over imagined disease trajectories and iteratively refines therapy proposals through survival-aware feedback, forming a closed-loop simulation-to-decision framework for treatment planning.
CLARITY achieves state-of-the-art performance in treatment planning and survival prediction across three cancer datasets, including two brain tumor cohorts (MU-Glioma-Post and zero-shot on UCSF-ALPTDG) and one breast cancer dataset (ISPY-2), demonstrating strong generalization across cancer types while consistently outperforming prior generative methods and medical-domain large language model baselines. 
\end{abstract}

\section{Introduction}
\label{sec:intro}

\begin{figure}[t]
    \centering
    \includegraphics[width=1\linewidth]{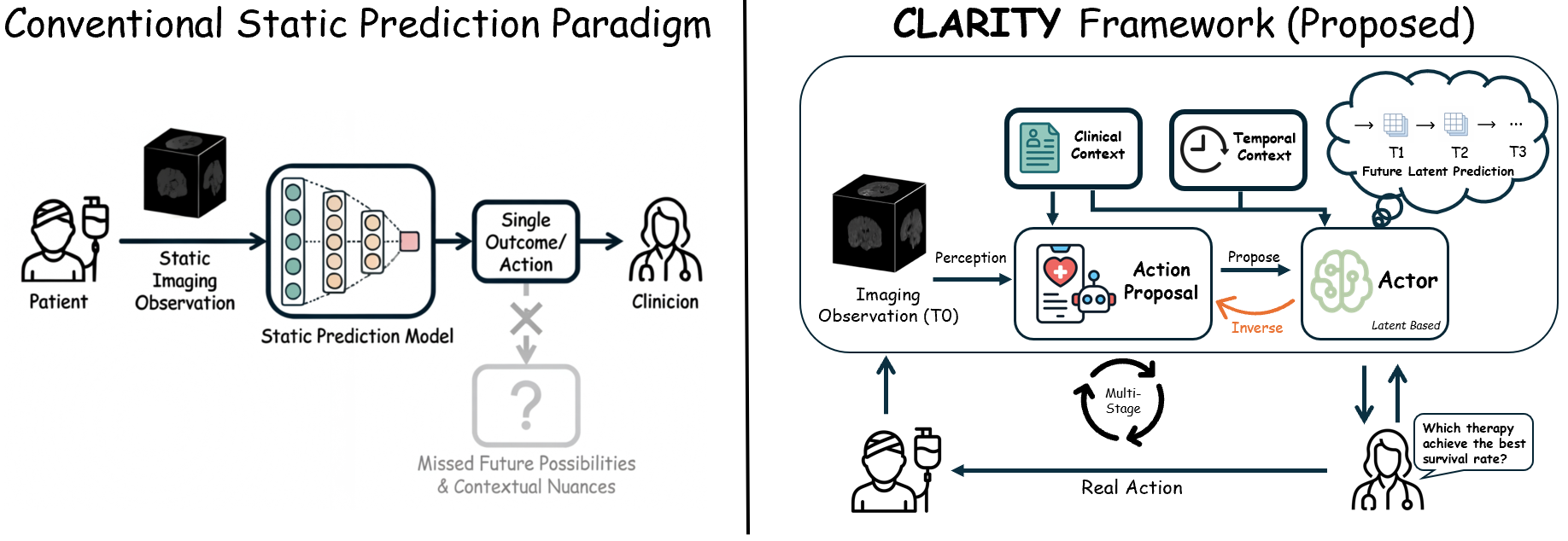}
    \caption{\textbf{Conceptual Overview of CLARITY}. Conventional clinical AI systems (\emph{left}) map imaging observations to a single predicted outcome, limiting their ability to evaluate alternative treatment strategies. In contrast, CLARITY ((\emph{right})) adopts a simulation-to-decision framework: a latent world model (Actor) simulates multiple ``what-if'' disease trajectories conditioned on clinical context, temporal intervals, and candidate therapies. 
These simulated trajectories are evaluated through survival-aware feedback, and the resulting signals iteratively refine therapy proposals (\textcolor{orange}{orange} arrow), enabling closed-loop optimization of treatment strategies.}
    \label{fig:intro}
    \vspace{-20pt}
\end{figure}

Clinical decision-making in oncology requires forecasting how a patient's disease evolves under treatment over time, a process characterized by substantial uncertainty. While modern AI systems achieve strong performance in static outcome prediction \cite{singhal2025toward, singhal2023large, amif, gloria, prior, mgca, mvcm,li2025fairfedmed,shi2025equitable,chen2024braixdet,luo2023harvard,liu2024translation}, they remain fundamentally limited in modeling treatment-conditioned disease trajectories across longitudinal follow-ups. However, real-world clinical decisions depend not only on predicting outcomes, but on anticipating how specific interventions reshape future physiological states. This makes treatment planning fundamentally a sequential decision-making problem, where clinicians must reason over potential disease trajectories under alternative interventions (e.g., continuing temozolomide therapy, introducing targeted agents, or escalating to salvage treatments), as illustrated in \cref{fig:intro}. Selecting an optimal treatment strategy is, therefore, an inherently difficult counterfactual task: clinicians must forecast how disease may evolve under therapies that were never observed; yet such prognostication remains highly uncertain and challenging, even for human experts~\cite{sagberg2022well,gwilliam2013prognosticating}.
Foundation models such as large language models (LLMs), despite impressive generalization capabilities \cite{hurst2024gpt4o, team2023gemini,al2025agentic}, are not designed for structured, temporally grounded disease forecasting. Dynamic clinical reasoning requires models that (i) represent physiological state transitions, (ii) incorporate heterogeneous patient-specific contexts, and (iii) support interpretable decision-making processes. Importantly, effective decision support requires evaluating counterfactual treatment strategies—predicting how the disease would evolve under interventions that have not yet been applied. 

World Models (WMs) offer a promising paradigm for such tasks. By learning structured latent dynamics, WMs enable forward simulation of future states and support planning over imagined trajectories \cite{bruce2024genie, hafner2025dreamerv3, v-jepa2, i-jepa}. While highly successful in robotics and control, their application to medicine remains \emph{limited}. Clinical data introduce unique challenges, including irregular time intervals, high inter-patient heterogeneity, multimodal conditioning, and stringent interpretability requirements.
Current medical generative approaches often forecast future clinical states by synthesizing raw images. However, effective clinical decision-making relies less on surface-level appearance generation and more on extracting robust visual features that accurately capture longitudinal disease progression. Because most current systems treat visual feature extraction and treatment optimization as completely disjointed stages, they lack an explicit mechanism to connect sequential visual state transitions with specific therapeutic interventions. Consequently, these models cannot support counterfactual reasoning about how different treatments may alter future disease dynamics. We therefore introduce CLARITY, a medical world model that directly links long-term visual forecasting with optimal treatment planning (\cref{fig:intro}). Unlike traditional predictive models that produce a single outcome estimate, CLARITY simulates multiple potential disease trajectories conditioned on candidate therapies and contextual factors. By modeling disease evolution directly within a latent space rather than synthesizing images, the framework captures continuous physiological transitions while enabling efficient long-horizon reasoning and counterfactual evaluation of treatment strategies.

A central design principle of CLARITY is explicit conditioning on \textbf{temporal and clinical contexts}. 
Continuous time intervals are encoded as embeddings to distinguish short-term response from long-term evolution, while multimodal patient attributes, including genomics, demographics, and therapeutic history are integrated to enable personalized trajectory modeling. This design allows the model to reason over irregular follow-up intervals and heterogeneous patient contexts commonly encountered in real-world clinical data.

Crucially, CLARITY bridges prediction and decision-making through a novel \textbf{Inverse Survival Evaluation}. 
Predicted latent rollouts are optimized under an entropy-regularized long-horizon objective and fed back into a therapy policy module, enabling iterative reassessment and refinement of treatment strategies. Instead of selecting therapies greedily from a single prediction, CLARITY performs planning over simulated disease trajectories, evaluating survival implications across candidate interventions before committing to an action.
This establishes a closed-loop prediction-to-decision framework that mirrors clinical reasoning: simulate potential trajectories, evaluate survival implications, and iteratively refine treatment policies. 

Overall, our contributions are threefold:
\begin{itemize}
    \item \textbf{Treatment-conditioned disease modeling.} We model longitudinal progression as continuous physiological transitions within a visual representation space, prioritizing biological consistency and enabling counterfactual modeling of treatment-induced disease dynamics.

    \item \textbf{Temporal and clinical conditioning.} We encode continuous time and multimodal patient context (e.g., genomics, demographics, prior therapies) to generate longitudinally coherent and individualized predictions.

    \item \textbf{Prediction-to-decision integration.} We introduce an inverse survival evaluation framework that transforms latent rollouts into adaptive and interpretable treatment recommendations.
\end{itemize}

\section{Related Works}
\label{sec:related works}

\noindent\textbf{World Models.}
The development of World Models has evolved along several paradigms. Generative-interactive models such as Genie \cite{bruce2024genie} simulate controllable environments from video, while latent planning frameworks such as DreamerV3 \cite{hafner2025dreamerv3} learn compact recurrent state-space models (RSSM \cite{doerr2018probabilistic_rssm}) for long-horizon control. More recently, non-reconstructive approaches such as V-JEPA \cite{v-jepa} and MuDreamer \cite{burchi2024mudreamer} predict representations directly in latent space without pixel reconstruction.
In medicine, however, world-model approaches remain limited. Existing methods typically rely on pixel-space generative modeling to predict future clinical states~\cite{MedWM,yue2025chexworld,yue2025echoworld,koju2025surgical,kraljevic2024foresight}. For example, MeWM \cite{MedWM} uses diffusion models to synthesize post-treatment tumor images. While these models produce visually plausible predictions, they focus on image synthesis rather than modeling treatment-conditioned disease dynamics for decision-making.
Our approach differs in three key aspects. First, existing methods typically predict a single future state, while our framework performs long-horizon trajectory simulation with time-conditioned latent dynamics. Second, diffusion-based image synthesis models do not naturally incorporate heterogeneous multimodal clinical context, whereas CLARITY explicitly conditions latent transitions on patient-specific clinical information for personalized forecasting and treatment planning. Third, prior medical generative models operate in pixel space and require reconstruction, whereas CLARITY models disease evolution directly through latent state transitions

\noindent\textbf{Survival Analysis.}
Traditional survival analysis relies on statistical models such as Cox regression \cite{cox_loss_cox1972regression} and Random Survival Forests \cite{ishwaran2008random}, which provide interpretable hazard estimation but assume proportional hazards and linear relations. Deep-learning extensions like DeepSurv \cite{katzman2018deepsurv}, DeepHit \cite{lee2018deephit}, and Deep Survival Machines \cite{nagpal2021deep} relax these assumptions, learning nonlinear risk mappings from high-dimensional clinical data. Recent multimodal frameworks \cite{gomaa2024comprehensive, mahmoudi2024multiparametric} further integrate imaging and omics features for personalized prognosis. Despite these advances, most survival models remain static predictors that estimate risk from observed patient states without modeling how disease trajectories evolve under different treatment decisions. Dynamic approaches such as Latent ODE~\cite{moon2022survlatent} and Dynamic DeepHit ~\cite{lee2019dynamic} introduce temporal continuity in survival modeling, yet they do not explicitly simulate treatment-conditioned disease trajectories or support long-horizon reasoning over alternative treatment strategies. In contrast, CLARITY integrates survival prediction with a latent world model that simulates treatment-conditioned disease evolution. By evaluating simulated trajectories under different therapies, the framework enables survival-aware planning over imagined futures rather than static outcome prediction.

\begin{figure*}[!t]
    \centering
    \includegraphics[width=1\linewidth]{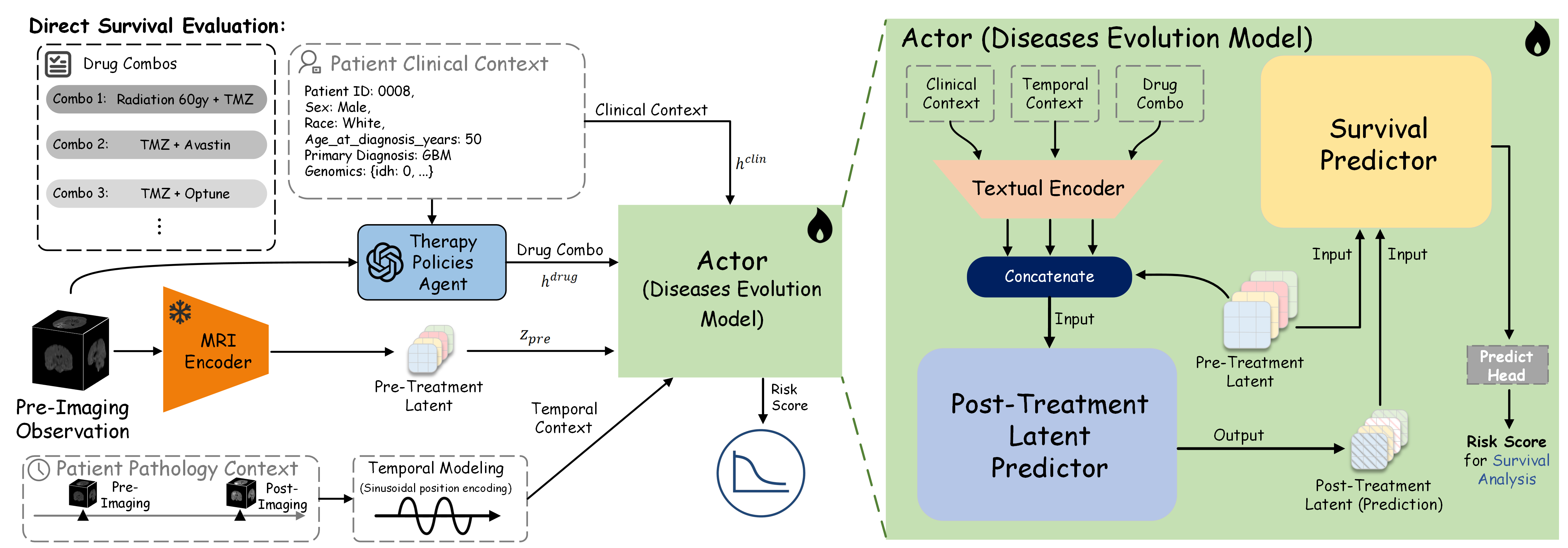}
    \caption{CLARITY's \textbf{Inference Pipeline} for \textbf{Direct Survival Evaluation}: A frozen MRI Encoder processes the pre-imaging observation to extract a pre-treatment latent representation. In parallel, the Therapy Policies Agent (\eg, GPT-5) takes the patient's clinical context to propose multiple candidate drug combos. The Actor module (Diseases Evolution Model) then sequentially evaluates each combo one-by-one, integrating the pre-treatment latent, clinical context, temporal context, and the specific drug combo to predict a final risk score for survival analysis.}
    \label{fig:direct_pipeline}
    \vspace{-10pt}
\end{figure*}
\vspace{-12pt}
\section{Method}
\label{sec:method}
\subsection{Overview}
\label{overview}
CLARITY consists of three main components: a parameter-efficient visual backbone (MRI Encoder), a Therapy Policy Agent implemented with a multimodal large language model (MLLM), and an Actor module (the Diseases Evolution Model). The pipeline first extracts a pre-treatment visual state $z_{\text{pre}}$ from the input MRI using a pre-trained vision backbone adapted via LoRA, which serves as the initial state for disease trajectory simulation. Conceptually, the Actor functions as a latent world model that simulates treatment-conditioned disease state transitions.

\begin{figure*}[!t]
    \centering
    \includegraphics[width=1\linewidth]{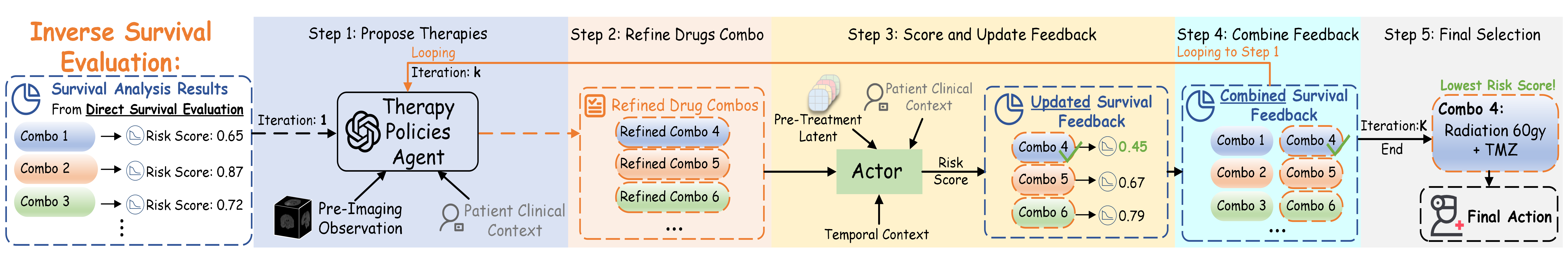}
    \caption{\textbf{CLARITY's Inverse Survival Evaluation}: This diagram illustrates the iterative prediction-to-decision feedback loop. Initial risk scores from Direct Survival Evaluation (\cref{fig:direct_pipeline}) are fed into the Therapy Policies Agent. The Agent then proposes updated drug combos, which the Actor Scores to generate risk estimates as the accumulated survival feedback. This process repeats, refining the therapy proposals, and after $K$ iterations, the policy with the Lowest Risk Score is selected as the Final Action.}
    \label{fig:inverse_evaluation}
\end{figure*}

CLARITY operates through two complementary inference modes. In \textbf{Direct Survival Evaluation} (\cref{fig:direct_pipeline}), the Therapy Policy Agent proposes candidate therapies based on the patient’s clinical context, and the Actor evaluates each option by integrating $z_{\text{pre}}$, clinical context, temporal context, and therapy embeddings to predict a survival risk score. In \textbf{Inverse Survival Evaluation} (\cref{fig:inverse_evaluation}), these scores are fed back to the agent to iteratively refine therapy proposals, enabling progressive search for treatments that minimize predicted risk.
During training, the Actor and LoRA adapters are optimized using paired pre- and post-treatment MRIs together with survival supervision to learn treatment-conditioned latent transitions. During inference, the MLLM is used only to generate candidate therapies, while the Actor evaluates their projected outcomes.
\vspace{-18pt}
\subsection{Therapy Policies Agent}
\label{sec:therapy policy}
In CLARITY, a multimodal large language model (MLLM) (\eg, GPT-5~\cite{singh2025openai}) serves as the \emph{Therapy Policy Agent} $\pi_{\text{MLLM}}$. This design allows the system to generate clinically valid treatment combinations while maintaining flexibility in exploring alternative therapy strategies. 
Conditioned on a high-level goal $g$ (\eg, ``minimize the predicted risk score''), the agent takes as input the pre-treatment MRI $x_0$ and clinical context $c_p$, and produces structured therapy actions that satisfy medical safety constraints~$\Omega$ 
(e.g., avoiding incompatible regimens such as co-administration of \textit{Bevacizumab} and \textit{Temozolomide}). 
Formally, the MLLM generates a candidate action set 
$\mathcal{A}^{(0)} = \{a_j^{(0)}\}_{j=1}^{M_0}$, 
where each $a_j^{(0)}$ represents a feasible treatment configuration. 
Each action $a$ contains various intervention components:
\begin{equation}
a = \{a_{\text{chemo}},\, a_{\text{radio}},\, a_{\text{brachy}},\, a_{\text{immuno}},\, a_{\text{add}}\},
\end{equation}
corresponding respectively to \textit{chemotherapy} (e.g., Temozolomide), 
\textit{external radiotherapy}, \textit{brachytherapy}, \textit{immunotherapy}, 
and \textit{additional} supportive strategies.
The generation process follows guideline-informed prompting templates to ensure clinical validity and parameter consistency 
(drug type, dose, and schedule).
Each generated therapy description is then encoded by a pretrained text encoder (\eg, \cite{sellergren2025medgemma}), 
yielding a dense embedding aligned with the Actor’s latent space:
\begin{equation}
h^{\text{drug}} = \text{Pool}(\Phi_{\text{text}}(a)) \in \mathbb{R}^{d},
\label{eq:drug_embedding}
\end{equation}
where $d$ denotes the embedding dimension of the latent space.

\subsection{Visual State Encoding}
\label{sec:visual_representation}
Let $x_0 \in \mathbb{R}^{H \times W \times D \times C}$ denote the pre-treatment 3D MRI scan. 
We employ a pre-trained 3D vision foundation model $\mathcal{F}_{\theta}(\cdot)$ 
(\eg, \cite{dong2025mri}) to encode anatomical and pathological features from the input volume.
To align the representation space with treatment-conditioned disease dynamics, we apply Low-Rank Adaptation (LoRA) \cite{hu2021lora} to the frozen backbone by injecting trainable low-rank matrices into its attention layers. This parameter-efficient adaptation allows the model to specialize to longitudinal disease modeling while preserving the general visual representations of the foundation model.
The adapted encoder maps the raw imaging observation $x_0$ to a latent visual state:
\begin{equation}
z_{\text{pre}} = \mathcal{F}_{\theta + \Delta \theta}(x_0) \in \mathbb{R}^{d_{pre}} ,
\end{equation}
where $\Delta \theta$ denotes the LoRA parameters and $d_{pre}$ is the latent representation dimension.
The resulting representation $z_{\text{pre}}$ serves as the initial disease state for the Actor module. 
During training, paired pre- and post-treatment MRIs allow the model to learn treatment-conditioned transitions between latent disease states, which are detailed in \cref{diseases evolution model}.
\vspace{-10pt}

\subsection{Clinical and Temporal Contexts}
\label{Clinical and Temporal Context}
The Actor receives not only the visual state $z_{\text{pre}}$ and therapy embeddings $h_{\text{drug}}$, but also \textbf{clinical and temporal conditioning signals} that modulate disease evolution dynamics.

\noindent\textbf{Clinical Context.}
For each patient $p$, we define a structured clinical profile $\mathcal{C}_p$ containing demographic attributes (e.g., age, sex), molecular biomarkers (e.g., \textit{IDH1/2}, \textit{ATRX}, \textit{1p19q} co-deletion, \textit{MGMT} methylation), and treatment-related indicators.
To obtain a unified representation of heterogeneous clinical variables, we serialize $\mathcal{C}_p$ into a textual prompt and encode it using a medical text encoder \cite{sellergren2025medgemma}.
The resulting embedding is then projected through a lightweight MLP to obtain the clinical representation
\[
h^{\text{clin}} = \mathrm{MLP}_{\text{clin}}\big(\Phi_{\text{text}}(\mathcal{C}_p)\big)
\in \mathbb{R}^{d_c},
\]
where $d_c$ denotes the clinical embedding dimension.
This representation conditions the latent disease dynamics, enabling the Actor to generate patient-specific trajectory predictions and survival estimates.

\noindent\textbf{Temporal Context.}
Disease progression occurs over irregular clinical follow-up intervals.  
Given observations acquired at times $t_{\text{pre}}$ and $t_{\text{post}}$, we encode the time gap $\Delta t = t_{\text{post}} - t_{\text{pre}}$ using sinusoidal embeddings:
\begin{equation}
   \gamma(\Delta t) = [\sin(\omega_i\Delta t), \cos(\omega_i\Delta t)]_{i=1}^{d_t/2},
   \label{eq:temporal encoding}
\end{equation}
where $\omega_i = 1/10000^{2i/d_t}$ and $d_t$ denotes the embedding dimension.
This continuous temporal encoding allows the Actor to model treatment-conditioned disease transitions across arbitrary follow-up intervals, providing explicit awareness of the elapsed time between observations.

\vspace{-10pt}
\subsection{Diseases Evolution}
\label{diseases evolution model}
\begin{wrapfigure}{t}{0.61\textwidth}
    \centering
    \vspace{-20pt}
    \includegraphics[width=\linewidth]{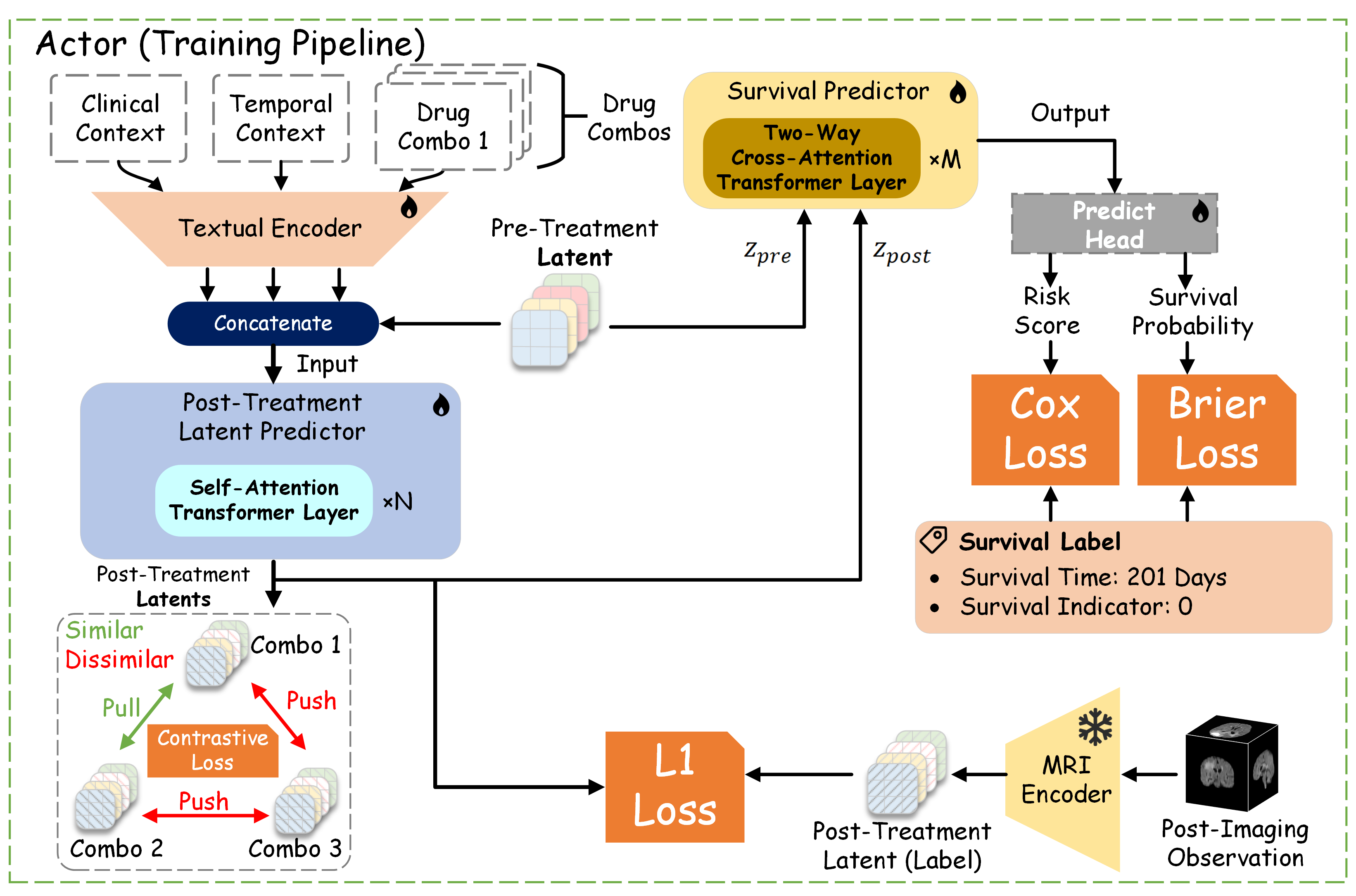}
    \caption{\textbf{Training Pipeline of the Actor}. The \textbf{Post-Treatment Latent Predictor} consists of a $N$-layer self-attention Transformer to forecast post-treatment latents. The \textbf{Survival Predictor} uses a $M$-layer two-way cross-attention Transformer to estimate risk score and survival rate.}
    \label{fig:training pipeline}
    \vspace{-15pt}
\end{wrapfigure}
As shown in \cref{fig:direct_pipeline}, we apply a Diseases Evolution Model as the Actor module, which models how a patient's disease state evolves under therapy and how this evolution relates to survival outcome. It consists of two submodules: a \textbf{Post-Treatment Latent Predictor}, which forecasts post-treatment latent representations, and a \textbf{Survival Predictor}, which estimates risk and survival probabilities from latent features. The detailed architectures of Post-Treatment Latent Predictor and Survival Predictor are shown in \cref{fig:training pipeline}. Conceptually, the Actor functions as a world model that predicts treatment-conditioned disease state transitions.

\noindent\textbf{Information Integration.}
To model treatment-conditioned state transitions, the Actor integrates four inputs into the Diseases Evolution Model. Specifically, the model receives pre-treatment latent $z_{\text{pre}}$, clinical context $h^{\text{clin}}$, temporal context $\gamma(\Delta t)$ and the drug combo $h^{\text{drug}}$. We concatenate these information input together as integrated representation, 
$[z_{\text{pre}},\, h^{\text{clin}},\, \gamma(\Delta t),\, h^{\text{drug}}]$.

\noindent\textbf{Post-Treatment Latent Predictor.}
The post-treatment latent predictor is optimized with a latent consistency loss and a soft contrastive regularization:
\begin{equation}
\mathcal{L}_{\text{pred}} = \lambda_1 \mathcal{L}_{\text{latent}} + \mathcal{L}_{\text{con}} .
\end{equation}
The state consistency term $\mathcal{L}_{\text{latent}}$ enforces consistency between the predicted latent state $\hat{z}_{\text{post}}$ and the ground-truth state $z_{\text{post}}$ using an $\ell_1$ loss:
\begin{equation}
\mathcal{L}_{\text{latent}} = \| \hat{z}_{\text{post}} - z_{\text{post}} \|_1
\end{equation}
where $d$ denotes the dimension of the latent space.
To structure the latent space, we adopt a soft-label contrastive objective $\mathcal{L}_{\text{con}}$. Rather than using binary similarity supervision, we compute pairwise treatment similarities $p_{ij}$ from the cosine similarities of treatment text embeddings and use them as soft targets. The predicted latent similarities $q_{ij}$ are then aligned to these targets through a temperature-scaled symmetric contrastive loss:
\begin{equation}
\mathcal{L}_{\text{con}} = -\sum_{i,j}^B (p_{ij} \log q_{ij} + q_{ij} \log p_{ij}) ,
\end{equation}
where $B$ is the batch size. This formulation encourages semantically similar treatments to induce nearby latent transitions, ensuring the latent space preserves the nuanced relationships of the clinical treatment space (e.g., therapies sharing similar mechanisms, such as alkylating chemotherapy agents, produce more similar disease transitions than fundamentally different interventions like radiation therapy).

\noindent\textbf{Survival Predictor.}
To connect latent disease evolution with clinical outcomes, we introduce a Survival Predictor that jointly processes the pre-treatment latent $z_{\text{pre}}$ and the predicted post-treatment latent $\hat{z}_{\text{post}}$. 
As illustrated in Fig.~\ref{fig:training pipeline}, we employ a novel \emph{bidirectional cross-attention module} to model interactions between disease state and treatment-induced transitions, enabling the model to capture how pre-treatment characteristics influence subsequent progression.
The aggregated representation produces two outputs: 
a one-year survival probability for calibrated outcome estimation and a continuous risk score for survival ranking. 
Following standard clinical survival modeling practice, we supervise these outputs using a Brier loss for calibration and a Cox partial likelihood loss for risk ordering~\cite{katzman2018deepsurv}. Our ablation experiments demonstrate that combining both losses yields more reliable survival prediction compared to single-loss variants.

\subsection{Inverse Survival Evaluation}
\label{inverse survival evaluation}

\noindent\textbf{Entropy-Regularized Long-Horizon Therapy Planning.}
Clinical treatment optimization is inherently sequential: therapy decisions influence future disease states, which in turn affect downstream survival outcomes. To capture this dependency, CLARITY formulates treatment selection as a long-horizon planning problem over imagined latent disease trajectories generated by the Actor. Rather than selecting a single intervention greedily, the framework evaluates entire treatment schedules while maintaining exploration to account for model uncertainty and clinical variability.
Formally, we model therapy planning as entropy-regularized trajectory optimization. Let $\mathbf{a}_{1:H}$ denote a treatment schedule over $H$ decision steps. The cumulative survival risk associated with the trajectory is defined as:
\begin{equation}
J(\mathbf{a}_{1:H}) 
= \sum_{t=1}^{H} \beta^{t-1} \hat r_t,
\end{equation}
where $\hat r_t$ is the predicted survival risk at step $t$ and $\beta$ is a temporal discount factor.
Instead of greedily minimizing $J$, we optimize a trajectory distribution $q$ under a maximum-entropy objective:
\begin{equation}
\max_{q} 
- \mathbb{E}_{q}[J(\mathbf{a}_{1:H})]
+ \tau \mathcal{H}(q),
\end{equation}
where $\mathcal{H}(q)$ denotes entropy and $\tau$ controls the exploration–exploitation trade-off.
In CLARITY, the trajectory distribution $q$ is implicitly parameterized by the Therapy Policy Agent implemented with an MLLM. At iteration $k$, the agent proposes a set of candidate therapy trajectories sampled from its conditional policy. Each trajectory is evaluated through long-horizon latent rollouts generated by the Actor, producing survival risk estimates.
These risk signals are incorporated into the next policy prompt as structured feedback, shifting probability mass toward lower-risk trajectories.

This loop effectively performs approximate entropy-regularized policy improvement: survival feedback reweights candidate trajectories while the stochastic generation capability of the language model maintains diversity in the search space. Over iterations, the distribution $q$ progressively concentrates around survival-optimal treatment strategies without prematurely collapsing exploration.

\vspace{-10pt}
\subsection{Arbitrary Policy-Conditioned Projection}
\label{arbitrary policy-conditioned projection}
CLARITY supports counterfactual ``what-if'' simulations by modeling dynamic treatment regimens as sequences of therapy actions $\{a_0, \dots, a_{K-1}\}$ executed at arbitrary discrete time points $\{t_0, \dots, t_K\}$. The Actor evaluates these policies by autoregressively generating a sequence of visual states.

Initializing the trajectory with the baseline visual state ($\hat{z}_0 = z_{\mathrm{pre}}$), at each step $k \in \{0, \dots, K-1\}$, the model applies the learned $z_{\mathrm{pre}} \rightarrow z_{\mathrm{post}}$ transition dynamics. Specifically, it integrates the current state $\hat{z}_k$ (acting as the pre-treatment anchor), clinical context $h^{\mathrm{clin}}$, time-gap $\phi(t_{k+1}-t_k)$, and therapy embedding $h^{\mathrm{drug}}_k$ to predict the subsequent state $\hat{z}_{k+1}$ (representing the post-treatment outcome). The Survival Predictor then maps this updated representation to stage-specific survival estimates $[\hat{p}_{1\mathrm{y}}^{(k+1)}, \hat{r}^{(k+1)}]$. 

This recursive rollout allows CLARITY to simulate disease trajectories under arbitrary policies \emph{without requiring paired imaging at intermediate steps}, supporting the inverse survival optimization process (qualitative examples in \cref{qualitative analysis}).

\vspace{-10pt}
\section{Experiment}
\vspace{-8pt}
\label{sec:experiment}

\subsection{Implementations and Datasets}
\vspace{-4pt}
\label{implementation}
For training and evaluation, we leverage two longitudinal brain tumor cohorts that provide both temporal imaging trajectories and treatment information, which are essential for modeling therapy-conditioned disease evolution in a world-model framework. 
The \textbf{MU-Glioma-Post} \cite{yaseen2025university} cohort contains 203 patients and 
654 MRI follow-ups with rich treatment logs and genomic annotations. 
We split the dataset at the \emph{patient level} to prevent temporal leakage, 
using 75\% of patients for training, 15\% for validation, and 15\% for testing. 
To ensure robustness, we further perform \emph{5-fold cross-validation} over the training-validation splits and report the mean performance across folds.
To evaluate generalization, we perform \textbf{zero-shot external validation} on the \textbf{UCSF-ALPTDG} \cite{data-ucsf-alptdg} brain tumor dataset, which contains 298 patients with longitudinal MRI scans and survival outcomes. The model is trained solely on MU-Glioma-Post and evaluated on UCSF-ALPTDG without fine-tuning.

Additionally, we evaluate CLARITY on the ISPY-2 breast cancer dataset \cite{wang2019spy}. This cohort contains 985 patients with longitudinal MRI scans across multiple treatment timepoints, alongside extensive molecular subtyping and neoadjuvant chemotherapy records. To demonstrate extensibility beyond neuro-oncology, the model is independently trained on this cohort to capture longitudinal therapeutic responses under neoadjuvant chemotherapy.

\subsection{Results on Treatment Exploration}
As shown in \cref{tab:comparison_expanded}, our CLARITY framework achieves state-of-the-art results, significantly outperforming all baselines across the evaluated datasets. On the MU-Glioma-Post benchmark, our approach achieves an F1-score of 57.1\%, a substantial 9.2\% absolute improvement over the strongest medical-specific baseline, Huatuo-Vision (46.4\%). The trend continues on UCSF-ALPTDG, where our model achieves 48.9\% F1, again outperforming the second-best method (44.1\%). 
These results highlight a key finding: while general-purpose models such as Claude-4.5 perform poorly when prompted directly (41.6\% F1), our framework successfully leverages the MLLM as part of a simulation-to-decision loop, substantially improving its effective performance. A critical comparison is with MeWM. Direct comparison with the original MeWM \cite{MedWM} was not feasible, as their in-house data is not publicly available and their work focuses on CT data, whereas ours targets multi-sequence MRI. To provide a fair baseline, we re-implemented a MeWM-style diffusion predictor (denoted as MeWM$^*$) using our MRI data. As shown in \cref{tab:comparison_expanded}, our approach (57.1\% F1) significantly outperforms this MeWM$^*$ baseline (43.6\% F1). This result validates our design choice to focus on \textbf{time-aware latent dynamics and survival-aware consistency} rather than diffusion-based reconstruction.
\begin{table}[t]
\centering
\small
\caption{\textbf{Quantitative comparison on the MU-Glioma-Post, UCSF-ALPTDG (zero-shot), and ISPY-2 datasets.}
MU-Glioma-Post and UCSF-ALPTDG are brain tumor datasets, while ISPY-2 is a breast cancer dataset.
The \textbf{best} and \underline{second-best} results are highlighted in boldface and underline, respectively.
$^*$ denotes re-implemented methods for fair comparison.}
\vspace{-10pt}
\label{tab:comparison_expanded}
\resizebox{\linewidth}{!}{%
\begin{tabular}{lcccccccccccc}
\toprule
\multirow{2}{*}{\textbf{Method}} & 
\multicolumn{4}{c}{\textbf{MU-Glioma-Post (\%)}} & 
\multicolumn{4}{c}{\textbf{UCSF-ALPTDG (\%)}} &
\multicolumn{4}{c}{\textbf{ISPY-2 (\%)}} \\
\cmidrule(lr){2-5} \cmidrule(lr){6-9} \cmidrule(lr){10-13}
 & Prec. & Rec. & F1 & Jac. & Prec. & Rec. & F1 & Jac. & Prec. & Rec. & F1 & Jac. \\
\midrule
\textit{General LLMs} \\
GPT-5 \cite{hurst2024gpt4o} 
& 49.3 & \underline{50.7} & \underline{50.0} & \underline{33.3} 
& 38.1 & 47.5 & 42.3 & 26.8
& 37.7 & 46.2 & 41.5 & 26.2 \\
Claude-4.5-Sonnet \cite{anthropic2024claude3.5} 
& 48.6 & 38.0 & 41.6 & 26.3
& \underline{45.3} & 38.6 & 41.7 & 26.3
& 42.8 & 36.5 & 39.4 & 24.5 \\
Qwen3-VL \cite{qwen3technicalreport,qwenvl-2.5}  
& 36.7 & 39.4 & 38.0 & 23.5
& 33.7 & 42.9 & 35.8 & 21.8
& 32.5 & 40.1 & 35.9 & 21.9 \\
\midrule
\textit{Medical models} \\
MedGPT \cite{MedicalGPT} 
& 41.6 & 42.1 & 41.9 & 26.5
& 36.7 & 46.3 & 40.9 & 25.7
& 38.2 & 43.5 & 40.7 & 25.6 \\
Huatuo-Vision \cite{chen2024huatuogpt}
& \underline{52.3} & 46.8 & 46.4 & 30.2
& 42.1 & \textbf{51.5} & \underline{44.1} & \underline{28.3}
& \underline{48.4} & \underline{49.1} & \underline{48.7} & \underline{32.1} \\
MeWM$^*$ \cite{MedWM} 
& 45.2 & 42.1 & 43.6 & 27.9
& 39.3 & \underline{48.2} & 43.3 & 27.6
& 41.5 & 45.3 & 43.3 & 27.7 \\
\midrule
\rowcolor{gray!10}
\textbf{Our Approach} 
& \textbf{61.3} & \textbf{53.5} & \textbf{57.1} & \textbf{39.9}
& \textbf{52.6} & 48.0 & \textbf{50.2} & \textbf{33.5}
& \textbf{56.2} & \textbf{51.8} & \textbf{53.9} & \textbf{36.8} \\
\bottomrule
\end{tabular}%
}

\end{table}
\begin{figure}[t]
    \centering
    \includegraphics[width=\linewidth]{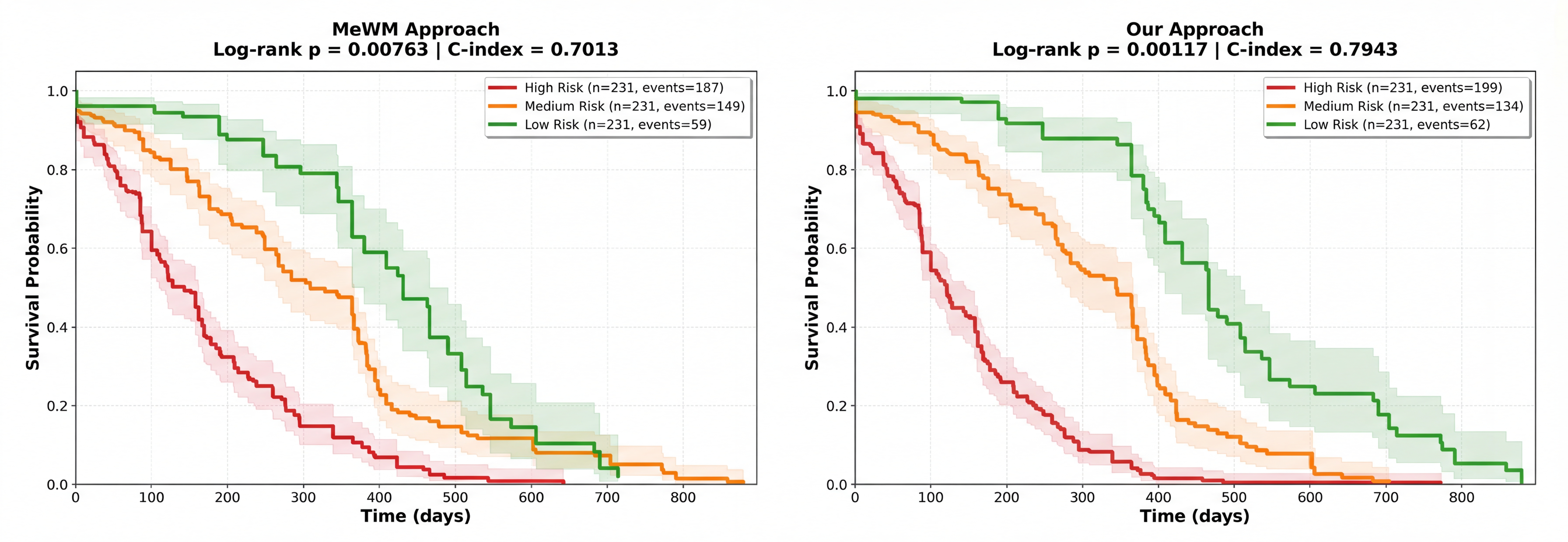}

    \caption{\textbf{Kaplan--Meier survival curves predicted by MeWM (left) and our method (right)} on MU-Glioma-Post. Our approach produces a much clearer separation across risk strata, reflected by a lower log-rank $p$-value of \textbf{0.0017} and a substantially higher C-index of \textbf{0.7943}. Shaded regions denote 95\% confidence intervals.}
    \label{fig:km_compare}
\end{figure}

\emph{Clinical prognostication itself is highly uncertain and challenging, even for experienced practitioners}. Prior studies report that neurosurgeons achieve roughly 40\% accuracy when predicting outcomes for high-grade glioma patients, while multidisciplinary oncology teams reach approximately 55\% accuracy in advanced cancer cohorts~\cite{sagberg2022well,gwilliam2013prognosticating}. These findings highlight the intrinsic difficulty of survival-oriented treatment planning. CLARITY achieves consistently higher predictive accuracy on the evaluated datasets, suggesting that trajectory-based simulation models can provide valuable decision-support signals for treatment planning. 
\vspace{-15pt}
\subsection{Survival Analysis}
\begin{wraptable}{r}{0.31\columnwidth}
\vspace{-25pt}
\centering
\caption{C-index performance of various methods.}
\setlength{\tabcolsep}{2.2pt}
\scriptsize
\begin{tabular}{lc}
\toprule
Method & C-index \\
\midrule
DeepSurv & 0.664 \\
RSF & 0.672 \\
SurvTRACE & 0.713\\
$MeWM^{*}$ & 0.701\\
CLARITY & \textbf{0.794$\pm$0.02} \\
\bottomrule
\end{tabular}
\vspace{-10pt}
\label{tab:surv_baselines}
\end{wraptable}

As shown in \cref{tab:surv_baselines}, CLARITY significantly outperforms standard survival modeling baselines (DeepSurv\cite{katzman2018deepsurv}, RSF\cite{ishwaran2008random}, SurvTRACE\cite{wang2022survtrace}), achieving a state-of-the-art C-index of $0.794$. 

Beyond quantitative metrics, \cref{fig:km_compare} demonstrates our model's superior discriminative capacity in Kaplan-Meier analysis compared to the diffusion-based MeWM baseline. CLARITY yields earlier and more sustained separation between high- and low-risk strata with minimal curve crossing, resulting in a substantially lower log-rank $p$-value ($0.0017$ vs.~$0.0763$). We attribute this robust stratification to our structured representation space, which avoids diffusion-induced artifacts and preserves structural consistency for reliable forecasting.


\vspace{-10pt}
\subsection{Analysis of Disease Decision Trajectory}
\label{qualitative analysis}
\begin{figure}[!t]
    \centering
    \includegraphics[width=0.8\linewidth]{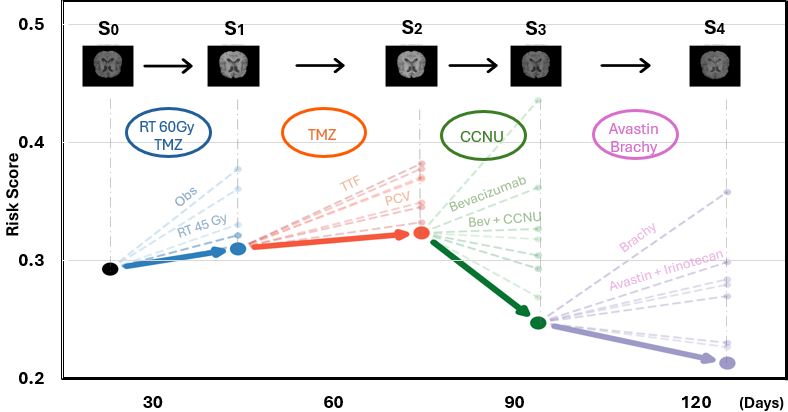}
    \caption{\textbf{Simulated Multi-Stage Decision Trajectories.} Stages $S_i$ denote MRI-anchored follow-ups. Dashed lines show \textbf{entropy-regularized exploration} of candidate therapies. The solid path highlights the \textbf{optimal sequence} minimizing cumulative risk $J$, adapting from standard-of-care (\eg, TMZ) to salvage regimens (\eg, CCNU, Avastin) based on state transitions. Boxed numbers denote 30-day risk scores.}
    \label{fig:trajectory}
\end{figure}

\cref{fig:trajectory} visualizes a predicted multi-stage decision trajectory. 
Each stage $S_i$ corresponds to an MRI observation and the associated latent state. 
At every stage, the model performs policy-conditioned state projection, generating multiple candidate branches (dashed lines) corresponding to different treatment actions.
The solid lane indicates the selected branch that achieves the lowest predicted risk score, with the corresponding therapy action labeled at each step. 
Through iterative selection, the model constructs a temporally consistent treatment lane (\eg, RT+TMZ $\rightarrow$ TMZ $\rightarrow$ CCNU $\rightarrow$ Avastin+Brachy) that minimizes the longitudinal risk trajectory.

\vspace{-10pt}

\subsection{Human Evaluation} 

For therapy recommendation, we use the recorded physician-administered regimen in MU-Glioma-Post as the reference regimen.  We report exact-match accuracy against the recorded regimen, while noting that the administered therapy represents an observed clinical decision rather than a unique optimal treatment. To complement this automatic metric, we further conducted a blinded expert assessment on 40 held-out cases. Two medical experts were presented with the patient context and two candidate regimens, corresponding to CLARITY's recommendation and the physician-administered regimen, with the source labels hidden. The experts selected the preferred regimen or marked the two as clinically equivalent. As shown in Table~\ref{tab:expert_eval}, CLARITY was judged equivalent/non-inferior in 60.0\% of cases and preferred over the physician regimen in 35.0\% of cases, while the recorded regimen was preferred in only 5.0\% of cases.
\begin{table}
\centering
\caption{\small Blinded expert assessment comparing CLARITY recommendations with recorded physician-administered regimens on 40 held-out MU-Glioma-Post cases.}
\label{tab:expert_eval}
\scriptsize
\setlength{\tabcolsep}{3pt}
\renewcommand{\arraystretch}{0.95}
\begin{tabular}{lc}
\toprule
Outcome & Cases (\%) \\
\midrule
Equivalent / non-inferior & \textbf{60.0} \\
CLARITY preferred & \textbf{35.0} \\
Physician preferred & 5.0 \\
\bottomrule
\end{tabular}
\vspace{-12pt}
\end{table}
These results indicate that CLARITY's recommendations are clinically acceptable in the vast majority of evaluated cases, with 95.0\% rated as either equivalent/non-inferior or preferable to the recorded regimen. This suggests that CLARITY does not simply imitate historical physician decisions, but can generate plausible alternative regimens that remain aligned with expert clinical reasoning. By combining guideline-informed candidate generation with survival-conditioned evaluation through the Actor, CLARITY produces patient-specific and outcome-aware therapy recommendations.

\subsection{Ablation Study}
\vspace{-3pt}
\subsubsection{Impact of CLARITY's Module.}
\cref{tab:combined_ablations} analyzes the contribution of each component. (1) \textbf{Latent vs. Diffusion:} Replacing diffusion synthesis (\# 1) with latent dynamics (\# 2) yields the largest gain (+8.8\% in F1), confirming that compact manifolds capture predictive structure better than stochastic pixel reconstruction. (2) \textbf{Clinical Context:} Integrating patient-specific priors (\# 3) enhances Recall, indicating that biological signals guide physiologically faithful simulations. (3) \textbf{Inverse Evaluation:} The feedback loop (\# 4 \& 5) drives performance to a peak F1-score of 57.1\% (\# 5). This validates that iteratively refining therapies via survival feedback is essential for optimized decision-making.

    

\noindent\textbf{Impact of Iteration Number.}
Tab.~\ref{tab:combined_ablations} investigates the impact of iteration number ($K$) on Inverse Survival Evaluation. We observe a steady performance gain as iterations increase, peaking at $K$=3 with a best F1-score of 57.1\%. This trajectory validates that iterative feedback enables our method to refine proposed therapies based on accumulated survival signals. However, the slight decline at $K$=4 indicates diminishing returns, suggesting that excessive iterations may introduce noise. Consequently, we adopt $K$=3 as the optimal trade-off.

\noindent\textbf{Impact of Loss Combination.}
Tab.~\ref{tab:combined_ablations} proves that richer supervision enhances predictive capacity. Adding the Brier score (\# 2) improves over the baseline (\# 1) by enforcing probability calibration (+0.7\%). Incorporating contrastive learning (\# 3) further sharpens latent transitions. Crucially, the soft-label variant (\# 4) yields the highest C-index (79.4\%), confirming that capturing nuanced treatment similarities stabilizes latent dynamics better than hard labels. 
\begin{table}[!t]
    \centering
    \scriptsize
    \caption{\textbf{Comprehensive Ablation Studies on MU-Glioma-Post.} (Top) Component analysis comparing architecture, context, and feedback. (Bottom Left) Impact of iteration $K$. (Bottom Right) Impact of loss combinations on the C-index.}
    \label{tab:combined_ablations}
    
    \setlength{\tabcolsep}{4pt}
    \begin{tabular}{@{}c|cccc|ccc@{}}
        \toprule
        \multirow{2}{*}{\#} &
        \multirow{2}{*}{\makecell{Diff.\\-based}} &
        \multirow{2}{*}{\makecell{Latent\\-based}} &
        \multirow{2}{*}{Context} &
        \multirow{2}{*}{\makecell{Feedback\\Iteration}} &
        \multicolumn{3}{c}{MU-Glioma-Post (\%)} \\
        \cmidrule(l){6-8}
        & & & & & Prec. & Rec. & F1 \\
        \midrule
        1 & \cmark & \xmark & \xmark & \xmark          & 47.8 & 39.8 & 43.6 \\
        2 & \xmark & \cmark & \xmark & \xmark          & 59.5 & 46.5 & 52.4 \\
        3 & \xmark & \cmark & \cmark & \xmark          & 58.5 & 48.1 & 52.8 \\
        4 & \xmark & \cmark & \xmark & \cmark~(K=3)    & 59.3 & 49.6 & 54.0 \\
        \rowcolor{gray!10}
        5 & \xmark & \cmark & \cmark & \cmark~(K=3)    & 61.3 & 53.5 & 57.1 \\
        \bottomrule
    \end{tabular}
    
    \vspace{8pt} 
    
    \setlength{\tabcolsep}{2.5pt} 
    \begin{tabular}{@{}cccc | l c@{}} 
        \toprule
        \multicolumn{4}{c|}{\textbf{Iteration Number ($K$)}} & \multicolumn{2}{c}{\textbf{Loss Combination}} \\
        \cmidrule(r{4pt}){1-4} \cmidrule(l{4pt}){5-6}
        $K$ & Prec. & Rec. & F1 & Loss Components & C-index (\%) \\
        \midrule
        1 & 59.7 & 48.1 & 53.1 & $\mathcal{L}_{\text{latent}} + \mathcal{L}_{\text{Cox}}$ & 76.1 \\
        2 & 59.3 & 48.6 & 53.4 & $\mathcal{L}_{\text{latent}} + \mathcal{L}_{\text{Brier}} + \mathcal{L}_{\text{Cox}}$ & 76.8 \\
        \rowcolor{gray!10} 
        3 & \textbf{61.3} & \textbf{53.5} & \textbf{57.1} & $\mathcal{L}_{\text{latent}} + \mathcal{L}_{\text{con}} \text{(Hard)} + \mathcal{L}_{\text{Brier}} + \mathcal{L}_{\text{Cox}}$ & 77.3 \\
        \rowcolor{gray!10} 
        4 & 57.2 & 53.6 & 55.4 & $\mathcal{L}_{\text{latent}} + \mathcal{L}_{\text{con}} \text{(Soft)} + \mathcal{L}_{\text{Brier}} + \mathcal{L}_{\text{Cox}}$ & \textbf{79.4} \\
        \bottomrule
    \end{tabular}
    \vspace{-10pt}
\end{table}

\noindent\textbf{Impact of Encoders.}
As shown in Tab.~\ref{tab:encoder_baseline}, the choice of visual encoder substantially influences survival prediction performance. 
Encoders pretrained on natural images (e.g., DINOv2) exhibit limited transferability to longitudinal MRI modeling, yielding the lowest C-index (0.665). 
In contrast, medical-domain pretraining (MedSigLIP) improves both ranking accuracy and probability calibration, suggesting that domain-aligned representations better capture clinically relevant patterns. 
Further gains are observed when leveraging MRI-specific encoders (brainIAC\cite{tak2026generalizable} and MRI-CORE\cite{dong2025mri}), which consistently outperform natural and generic medical models. 
In particular, MRI-CORE achieves the best C-index (0.794) and lowest Brier score (0.162), indicating that modality-aligned pretraining combined with lightweight adaptation (LoRA) provides more discriminative and stable latent representations for downstream survival reasoning.
\begin{table}[!t]
\centering
\scriptsize
\caption{All methods share the same world model and training protocol.
We report mean $\pm$ std over 5 random seeds.
Higher C-index are better; lower Brier score is better.}

\label{tab:encoder_baseline}
\setlength{\tabcolsep}{3.5pt}
\renewcommand{\arraystretch}{1.0}

\begin{tabular}{lccccc}
\toprule
\textbf{Encoder} 
& \textbf{Pretrain} 
& \textbf{Frozen} 
& \textbf{C-index} 
& \textbf{Brier} \\
\midrule
DINOv2  & Natural (SSL) & Yes  & 0.665$\pm$0.020 & 0.198$\pm$0.008 \\
MedSigLIP & Medical & Yes & 0.705$\pm$0.016 & 0.176$\pm$0.006 \\
brainIAC & MRI & LoRA & \underline{0.786$\pm$0.012} & \underline{0.168$\pm$0.004} \\
MRI-CORE & MRI & LoRA & \textbf{0.794$\pm$0.010} & \textbf{0.162$\pm$0.003} \\
\bottomrule
\end{tabular}

\end{table}

\vspace{-10pt}
\noindent\textbf{Comparison between Diffusion and Latent Representation.}
Tab.~\ref{tab:efficiency_drift} highlights the dual advantages of our latent-based dynamics in both computational efficiency and feature fidelity. First, the diffusion-based approach is prohibitively expensive, requiring up to 61.3 TFLOPs and 38.6 seconds for a \textit{single} simulation. In contrast, our latent-based predictor reduces computation to 4.21 TFLOPs (an approximate 9--15$\times$ reduction) and achieves sub-second inference (0.341s). This efficiency gain is essential, making our Inverse Survival Evaluation computationally feasible even in a multi-iteration setting (\ie, $K=3$). Second, we quantify the fidelity loss incurred by the diffusion model's pixel-space detour (generation followed by re-encoding). Keeping the encoder and survival head identical to ensure a fair comparison, we measure the cosine distance to the ground-truth post-treatment latent. As shown in Tab.~\ref{tab:efficiency_drift}, diffusion-based reconstruction leads to nearly $2\times$ higher representation drift. This indicates that the intermediate pixel-generation step introduces substantial noise, whereas our direct latent prediction maintains the stability required for downstream survival evaluation.
\begin{table}[!t]
    \centering
    \scriptsize
    \setlength{\tabcolsep}{6pt}
    \caption{Comparison of efficiency and drift between diffusion and latent methods.}
    \vspace{-10pt}
    \label{tab:efficiency_drift}
    \begin{tabular}{lcccc}
        \toprule
        \multirow{2}{*}{Method} & \multicolumn{2}{c}{Efficiency} & \multicolumn{2}{c}{Representation Drift} \\
        \cmidrule(lr){2-3} \cmidrule(lr){4-5}
        & FLOPs (T) $\downarrow$ & Time (s) $\downarrow$ & Cosine Dist. $\downarrow$ & Ratio $\uparrow$ \\
        \midrule
        Diffusion-based (1000 steps) & 61.3 & 38.6 & \multirow{2}{*}{0.0174} & \multirow{2}{*}{-} \\
        Diffusion-based (500 steps)  & 39.5 & 19.7 & & \\
        \rowcolor{gray!10} Latent-based (Ours) & \textbf{4.21} & \textbf{0.341} & \textbf{0.0084} & $\mathbf{1.9\times}$ \\
        \bottomrule
    \end{tabular}
    \vspace{-10pt}
\end{table}

\vspace{-10pt}
\section{Conclusion}
\vspace{-10pt}
\label{sec:conclusion}
We introduced CLARITY, a medical world model addressing critical limitations in prior works. By forecasting disease evolution in a \textbf{latent space}, we avoid the stochasticity and high computational cost of diffusion-based models. CLARITY is the first to explicitly integrate \textbf{temporal and clinical contexts} with an \textbf{Inverse Survival Evaluation}. Experiment proves that our feedback loop is critical for optimized decisions, with results outperforming all baselines. Therefore, CLARITY represents a significant step towards computationally feasible and personalized treatment planning in oncology.

{\small
\bibliographystyle{splncs04}
\bibliography{main}
}

\section{Additional Implementation Details}
\label{sec:additional_implementation_details}
CLARITY consists of four major components: an MRI encoder, a Therapy Policy Agent, a text encoder, and an Actor module.

\begin{itemize}
    \item \textbf{MRI Encoder:} For visual state encoding, we utilize the pre-trained MRI-CORE foundation model \cite{dong2025mri}. Specifically, we adopt its 12-layer Vision Transformer Base (ViT-B) architecture, which extracts features from the input 3D multi-sequence MRI scans and maps them into a compact latent space with a dimension of $d=768$. To efficiently adapt this backbone for longitudinal trajectory modeling, we apply Low-Rank Adaptation (LoRA)\cite{hu2021lora} to its attention layers while keeping the original pre-trained weights frozen.
    \item \textbf{Therapy Policy Agent:} We employ GPT-5 to generate candidate actions.
    \item \textbf{Text Encoder:} We use MedGemma\cite{sellergren2025medgemma}, which is fine-tuned via 4-bit quantization and LoRA to obtain representations of heterogeneous clinical variables.
    \item \textbf{Actor Module:} The Diseases Evolution Model applies a Transformer with $N=4$ self-attention layers as its Post-Treatment Latent Predictor, and a Transformer with $M=4$ two-way cross-attention layers as its Survival Predictor.
\end{itemize}
For the training strategy, we use the AdamW optimizer.

\section{Detailed Loss Formulations}
\label{sec:detailed_loss}
We optimize the Actor module using a composite objective function that balances latent consistency, structured representation learning, and survival analysis calibration. The total loss is defined as:

\begin{equation}
    \mathcal{L}_{total} = \lambda_1\mathcal{L}_{latent} + \lambda_2\mathcal{L}_{con} + \lambda_3\mathcal{L}_{Brier} + \lambda_4\mathcal{L}_{Cox}
\end{equation}

where we empirically set the coefficients to $\lambda_1=5$, $\lambda_2=1$, $\lambda_3=1$, and $\lambda_4=1$.

The Brier score loss ($\mathcal{L}_{Brier}$) enforces probability calibration for the one-year survival estimate:

\begin{equation}
    \mathcal{L}_{Brier} = \frac{1}{B}\sum_{i=1}^{B}(\hat{p}_i - y_i)^2
\end{equation}

where $\hat{p}_i$ is the predicted one-year survival probability, $y_i \in \{0, 1\}$ is the ground-truth survival indicator, and $B$ is the batch size.

The Cox partial likelihood loss ($\mathcal{L}_{Cox}$) \cite{cox1972regression} maintains the correct risk ordering across patient trajectories:

\begin{equation}
    \mathcal{L}_{Cox} = -\frac{1}{N_{E}}\sum_{i:E_i=1}\left(\hat{r}_i - \log\sum_{j\in R(T_i)}\exp(\hat{r}_j)\right)
\end{equation}

where $E_i$ indicates if an event occurred, $N_{E}$ is the total number of observed events, $\hat{r}_i$ is the predicted continuous risk score, and $R(T_i)$ is the risk set of patients who survived at least up to time $T_i$.

\section{Interpretability of Latent Disease Dynamics}
\label{sec:interpretability}
Unlike prior medical world models that rely on pixel-space synthesis for forecasting, CLARITY models longitudinal disease evolution directly through continuous latent state transitions to ensure computational efficiency and structural consistency. However, to demonstrate that our predicted latent states ($\hat{z}_{post}$) represent interpretable, biologically meaningful disease dynamics rather than abstract noise, we trained an auxiliary diffusion-based decoder purely as an analytical probe. 

\begin{figure}[ht]
    \centering
    \begin{subfigure}[b]{0.23\linewidth}
        \centering
        \includegraphics[width=\linewidth]{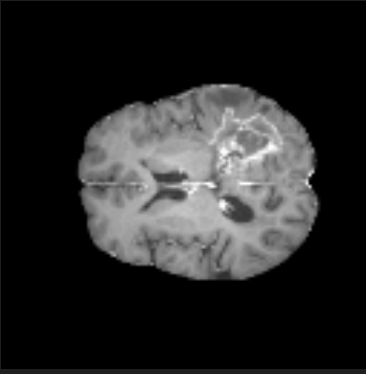}
    \end{subfigure}\hfill
    \begin{subfigure}[b]{0.23\linewidth}
        \centering
        \includegraphics[width=\linewidth]{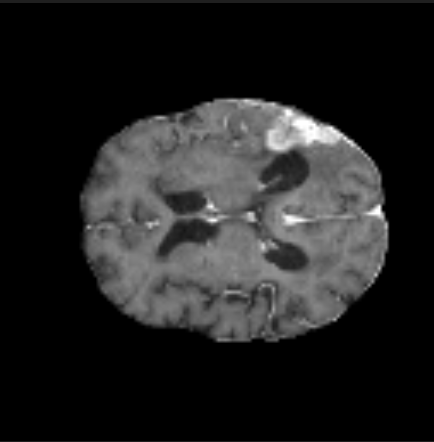}
    \end{subfigure}\hfill
    \begin{subfigure}[b]{0.23\linewidth}
        \centering
        \includegraphics[width=\linewidth]{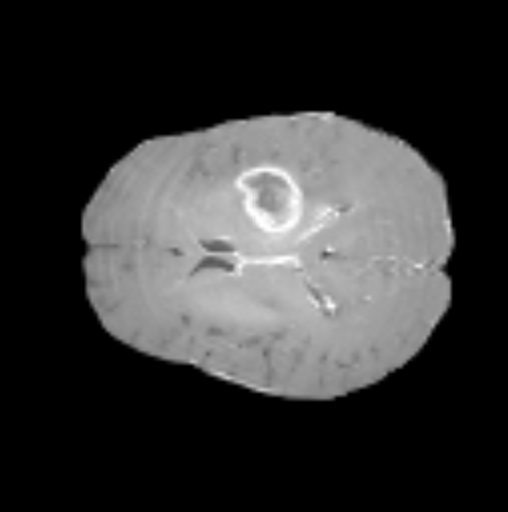}
    \end{subfigure}\hfill
    \begin{subfigure}[b]{0.23\linewidth}
        \centering
        \includegraphics[width=\linewidth]{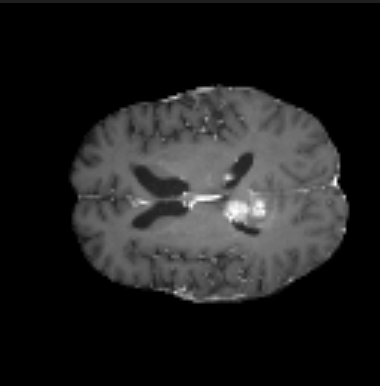}
    \end{subfigure}
    
    \vspace{0.2em} 
    
    \begin{subfigure}[b]{0.23\linewidth}
        \centering
        \includegraphics[width=\linewidth]{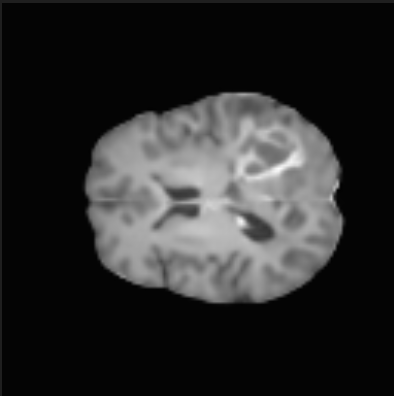}
        \caption{Case 1}
    \end{subfigure}\hfill
    \begin{subfigure}[b]{0.23\linewidth}
        \centering
        \includegraphics[width=\linewidth]{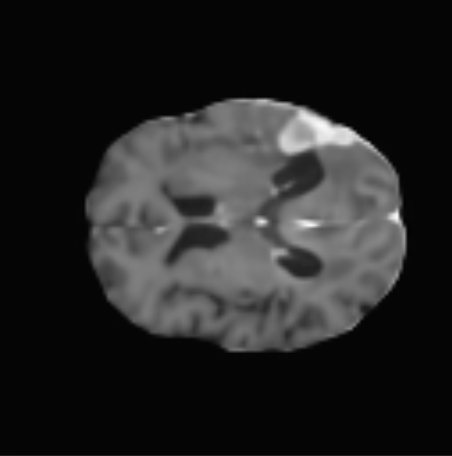}
        \caption{Case 2}
    \end{subfigure}\hfill
    \begin{subfigure}[b]{0.23\linewidth}
        \centering
        \includegraphics[width=\linewidth]{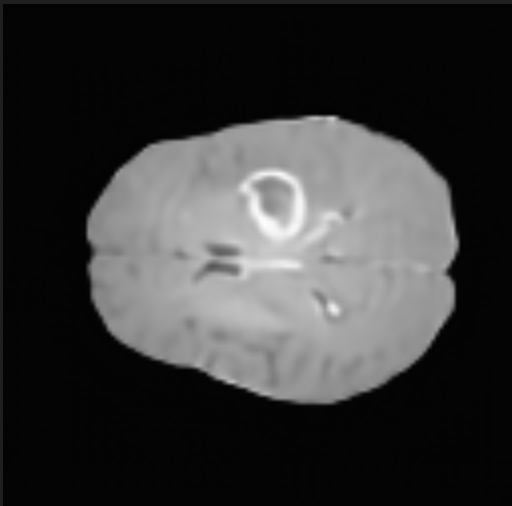}
        \caption{Case 3}
    \end{subfigure}\hfill
    \begin{subfigure}[b]{0.23\linewidth}
        \centering
        \includegraphics[width=\linewidth]{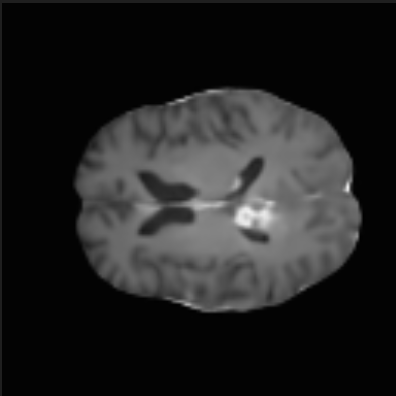}
        \caption{Case 4}
    \end{subfigure}
    
    \caption{\textbf{Qualitative visualization probing the interpretability of predicted latent states.} The top row displays the ground-truth post-treatment MRIs, while the bottom row shows the semantic information decoded from the predicted latent vectors ($\hat{z}_{post}$). Because the compact latent space acts as a semantic bottleneck, it naturally abstracts away stochastic, high-frequency textural noise (resulting in visually smoother images). However, it faithfully preserves the highly interpretable macroscopic anatomical structures and clinically relevant pathological changes.}
    \label{fig:latent_recon_visual}
\end{figure}

This decoder maps the Actor's predicted post-treatment latents back into the MRI pixel space, enabling qualitative and quantitative inspection of the semantic information captured by our latent dynamics. \cref{tab:latent_recon} details the perceptual and structural fidelity of these decoded latent representations compared against the ground-truth post-treatment MRIs.

\begin{table}[ht]
\centering
\scriptsize
\caption{Quantitative evaluation of the information encoded in the predicted latent states, assessed via decoded MRI reconstructions on the MU-Glioma-Post dataset.}
\label{tab:latent_recon}
\vspace{0.3em}
\begin{tabular}{lcc}
\toprule
\textbf{Metric} & \textbf{Mean} & \textbf{Std Dev} \\
\midrule
\textbf{PSNR (dB) $\uparrow$} & 27.91 & 1.12 \\
\textbf{SSIM $\uparrow$} & 0.890 & 0.024 \\
\textbf{LPIPS $\downarrow$} & 0.672 & 0.018 \\
\textbf{FID $\downarrow$} & 0.710 & 0.020 \\
\bottomrule
\end{tabular}
\end{table}

The quantitative results in \cref{tab:latent_recon} robustly validate the interpretability of our latent space:

\begin{itemize}
    \item \textbf{Macro-Anatomy Preservation (PSNR \& SSIM):} High PSNR (27.91 dB) and SSIM (0.890) confirm that the predicted latent state $\hat{z}_{post}$ securely encodes the patient's global brain structure. As visible in \cref{fig:latent_recon_visual}, despite textual smoothing, the model successfully updates pathological features without corrupting the underlying spatial boundaries or patient identity, proving the latent representation is anatomically grounded.
    
    \item \textbf{Semantic Disease Evolution (FID \& LPIPS):} An FID of 0.710 demonstrates high distributional realism. Concurrently, the LPIPS of 0.672 reflects a deliberate encoding of biologically meaningful, therapy-induced morphological changes (e.g., variations in tumor volume or mass effect) rather than stochastic pixel noise. This proves the latent space captures an interpretable, abstracted trajectory of the disease rather than simply performing an identity mapping of the pre-treatment scan.
\end{itemize}

\section{Clinical Case: Simulated Treatment Trajectories}
\label{sec:case_studies}
\begin{figure}[htbp]
    \centering
    \begin{tcolorbox}[
        enhanced, 
        colback=uibg,           
        colframe=uiheader,      
        colbacktitle=uiheader,  
        coltitle=white,         
        fonttitle=\sffamily\bfseries\normalsize, 
        title={Case 1: Glioblastoma, WHO Grade 4 \hfill Patient ID: 0014}, 
        arc=3pt,                
        boxrule=0.8pt,
        left=8pt, right=8pt, top=4pt, bottom=4pt,
    ]
    
    \noindent\textcolor{uitext}{\sffamily\textbf{PATIENT PROFILE}} \hfill 
    {\scriptsize \sffamily \textbf{Demographics:} 53yo Female ~|~ \textbf{Surgery:} Initial surgery at day -2}\\[0.3em]
    \genetag[red!15]{IDH1 Wild-type} \hspace{1pt}
    \genetag[red!15]{MGMT Unmethylated} \hspace{1pt}
    \genetag[uiblue!15]{EGFR Amplified} \hspace{1pt}
    \genetag[uigreen!15]{1p/19q Intact} \hspace{1pt}
    \genetag[gray!20]{ATRX Altered}
    
    \vspace{0.5em}
    \noindent\textcolor{black!15}{\rule{\linewidth}{1pt}} 
    \vspace{0.3em}
    
    \noindent\textcolor{uitext}{\sffamily\textbf{CLINICAL TIMELINE (GROUND TRUTH)}}\\[0.3em]
    \begin{minipage}[c]{0.72\linewidth}
        {\footnotesize \sffamily
        \renewcommand{\arraystretch}{1.65} 
        \begin{tabular}{@{}p{0.22\linewidth} p{0.75\linewidth}@{}}
        \textcolor{uiblue}{\textbf{Day 0}} & \textbf{Diagnosis:} Confirmed GBM \\
        \textcolor{uiblue}{\textbf{Day 19--61}} & \textbf{Concurrent Tx:} TMZ + RT (60 Gy/30 fx) \\
        \textcolor{uiblue}{\textbf{Day 113--566}} & \textbf{Adjuvant Tx:} Maintenance TMZ (q28 days $\times$ 17 cycles) \\
        \textcolor{uired}{\textbf{Day 456--566}} & \textbf{Salvage Tx:} Avastin initiated (q14 days $\times$ 8 cycles) \\
        \textcolor{uired}{\textbf{Day 552}} & \textbf{Progression:} First progression confirmed \\
        \multicolumn{2}{@{}l}{\textcolor{black!50}{\textit{MRI Follow-ups: Days 84, 126, 246, 316, 351, 442}}} \\
        \end{tabular}
        }
    \end{minipage}%
    \hfill
    \begin{minipage}[c]{0.25\linewidth}
        \centering
        \includegraphics[width=0.85\linewidth]{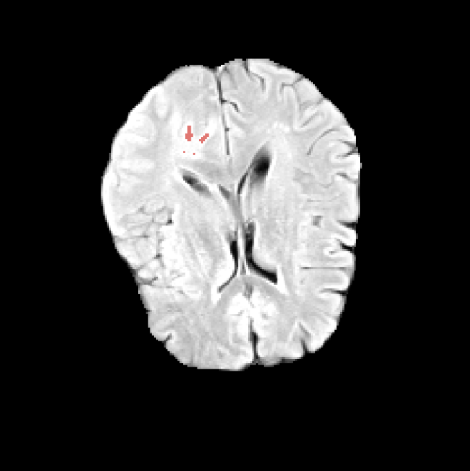} \\[-0.1em]
        {\tiny \sffamily \textcolor{uitext}{Day 0: Pre-treatment}} \\[0.5em]
        \includegraphics[width=0.85\linewidth]{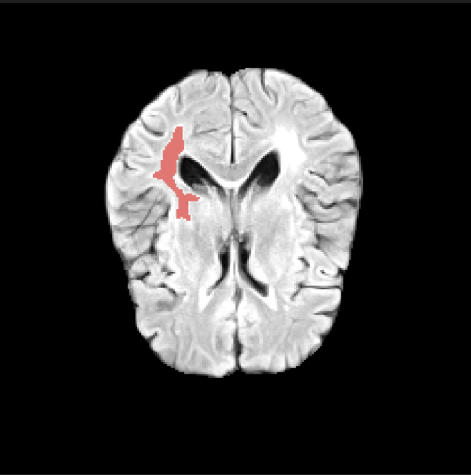} \\[-0.1em]
        {\tiny \sffamily \textcolor{uired}{Day 552: Progression}}
    \end{minipage}
    
    \vspace{0.6em}
    \noindent\textcolor{black!15}{\rule{\linewidth}{1pt}} 
    \vspace{0.3em}
    
    \noindent\textcolor{uitext}{\sffamily\textbf{PREDICTION \& F1 ALIGNMENT ANALYSIS}}\\[0.3em]
    {\scriptsize \sffamily 
    \renewcommand{\arraystretch}{1.5} 
    \begin{tabular}{@{}p{0.18\linewidth} p{0.37\linewidth} p{0.45\linewidth}@{}}
    \toprule
    \textbf{Decision Phase} & \textbf{Ground Truth (Target)} & \textbf{CLARITY (Predicted)} \\
    \midrule
    
    \textbf{Initial Strategy} & 
    TMZ ($75\text{ mg/m}^2/\text{d}$) \newline 
    RT (60 Gy / 30 fx) & 
    TMZ ($75\text{ mg/m}^2/\text{d}$) \textcolor{uigreen}{\textbf{[\checkmark Match]}} \newline 
    RT (60 Gy / 30 fx) \textcolor{uigreen}{\textbf{[\checkmark Match]}} \\
    
    \textbf{Adjuvant Tx} & 
    TMZ (q28d $\times$ 17 cycles) & 
    TMZ (q28d $\times$ 12 cycles) \textcolor{orange}{\textbf{[Partial]}} \newline 
    Optune TTF \textcolor{uiblue}{\textbf{[Add-on]}} \\
    
    \textbf{Salvage Tx \newline (Recurrence)} & 
    Avastin (q14d $\times$ 8 cycles) & 
    Avastin (q14d $\times$ 6 cycles) \textcolor{orange}{\textbf{[\checkmark Partial]}} \\
    
    \bottomrule
    \end{tabular}
    }
    \end{tcolorbox}
    \caption{\textbf{Structured Clinical Case Study (Patient 0014).} The card visualizes the patient's baseline genomic markers and the ground-truth clinical timeline (top/middle) alongside corresponding pre-treatment and progression MRI scans. In the prediction analysis (bottom), CLARITY demonstrates high clinical alignment by accurately reproducing the initial chemoradiotherapy regimen and dynamically recommending an earlier transition to Avastin salvage therapy, reflecting appropriate reasoning for an MGMT-unmethylated (TMZ-resistant) tumor phenotype.}
    \label{fig:case_study_0014}
\end{figure}

\cref{fig:case_study_0014} illustrates CLARITY's decision-making for a patient with MGMT GBM, a phenotype typically resistant to standard Temozolomide (TMZ) therapy. 
While matching all ground-truth core modalities, CLARITY demonstrates superior clinical context-awareness by recommending an earlier cessation of adjuvant TMZ (12 vs. 17 cycles) and a timely transition to Avastin-based salvage therapy. 
\section{Therapy Policy Agent Prompting and Safety}
\label{sec:prompting_safety}
The multimodal large language model utilizes guideline-informed prompting templates to ensure clinical validity and parameter consistency. Conditioned on the goal to minimize the predicted risk score, the prompt structure provides the agent with the patient's profile and enforces a rigid output schema mapping to the action components.

The constraint set $\Omega$ enforces clinical validity at all stages:
\begin{enumerate}
    \item The policy cannot propose incompatible drug pairs (e.g., co-administration of Bevacizumab and Temozolomide outside of specific salvage protocols).
    \item Dose and cycle proposals generated from the distribution $q$ are clipped to established guideline ranges (e.g., radiation dose is constrained to $40-60$ Gy).
    \item History-aware rules prevent conflicts with prior lines of therapy (e.g., preventing the repetition of identical modalities unless recurrence is explicitly suggested).
\end{enumerate}

This feedback-driven, constraint-aware loop ensures simulated outcomes mirror safe, iterative clinical plan refinement.

\subsection{System Prompt Template}
To ensure reproducibility and demonstrate how clinical constraints ($\Omega$) are structurally enforced during the generative process, we provide the exact system prompt utilized by the Therapy Policy Agent. The prompt explicitly defines the JSON schema, output ranges, and context-mapping rules for the Inverse Survival Evaluation. \textbf{The complete prompt template is visualized in \cref{fig:system_prompt}.}

\noindent
\section{Limitations}
\label{sec:limitations}
We acknowledge several limitations in our current work:
\begin{itemize}
    \item \textbf{Generalizability:} Our primary brain tumor model is trained on specific glioma cohorts (MU-Glioma-Post ). Its direct performance across unseen clinical centers or significantly altered imaging protocols requires further multi-institutional validation.
    
    \item \textbf{Domain Specificity:} While the framework successfully models both brain gliomas and breast cancer (I-SPY2 \cite{wang2019spy}), it currently relies heavily on multi-sequence MRI. Adapting it to other imaging modalities, such as CT, would require substantial representation realignment and re-training.
    
    \item \textbf{Therapy Agent Constraints:} The policy agent is constrained by a predefined set of medical rules ($\Omega$). An incomplete $\Omega$ could lead to sub-optimal therapy proposals or fail to capture novel, off-label clinical trial interventions.
    
\end{itemize}

\begin{figure}[htbp] 
    \centering
    \begin{tcolorbox}[enhanced, colback=white, colframe=black, boxrule=0.8pt, sharp corners, left=10pt, right=10pt, top=10pt, bottom=10pt]
    \small \sffamily 

    You are a neuro-oncologist AI proposing POST treatment actions for glioblastoma.\\
    ---
    
    \textcolor{teal}{\textbf{\#\#\# * Task Description *}}\\
    1. Analyze the patient's pre-treatment clinical context and generated feedback.\\
    2. Output potential POST treatment therapy sequences. The candidates must be output in JSON format, adhering strictly to the schema and clinical safety constraints.\\
    ---
    
    \textcolor{red}{\textbf{\#\#\# * Constraints \& Safety Rules *}}\\
    \textcolor{blue}{\textbf{\#\#\#\# * Dosage \& Cycles *}}\\
    - radiation: \texttt{dose\_gy} $\in$ [40, 60], \texttt{fractions} $\in$ [10, 33]\\
    - drugs: \texttt{cycle\_length\_days} $\in$ [14, 56], \texttt{num\_cycles} $\in$ [1, 12]\\
    - Generate diverse candidates covering a spectrum of post-treatment intensities (e.g., supportive care, standard maintenance, aggressive combination therapy). At least one candidate must represent "no active therapy / observation".\\
    - Omit empty arrays. No-post outputs are valid (e.g., \texttt{"actions": \{\}}).\\
    
    \textcolor{blue}{\textbf{\#\#\#\# * Clinical Mapping Rules *}}\\
    - Maintain temporal and clinical consistency with the PRE treatment payload.\\
    - Transitions from active concurrent therapies to adjuvant maintenance phases should strictly follow standard neuro-oncology guidelines.\\
    - Never repeat identical radiation or chemotherapy agents that already appear in the PRE payload, unless explicitly indicating recurrence.\\
    - \textbf{Feedback Loop:} If the user payload includes a \texttt{"feedback"} field, treat it as a survival risk critique from the world model (lower total scores indicate better survival). You \textbf{MUST} adjust subsequent proposals to minimize this risk score.\\
    ---
    
    \textcolor{orange}{\textbf{\#\#\# * Example Schema *}}\\
    \textbf{**Example Output**}
    
    \texttt{\{}\\
    \hspace*{1em}\texttt{"candidates": [}\\
    \hspace*{2em}\texttt{\{}\\
    \hspace*{3em}\texttt{"post": \{}\\
    \hspace*{4em}\texttt{"tp": "TP\_post",}\\
    \hspace*{4em}\texttt{"actions": \{}\\
    \hspace*{5em}\textcolor{purple}{\texttt{"radiation"}}\texttt{: [\{"dose\_gy": 60, "fractions": 30\}],}\\
    \hspace*{5em}\textcolor{purple}{\texttt{"chemotherapy"}}\texttt{: [\{"agent": "Temozolomide", "cycle\_length\_days": 28, "num\_cycles": 6\}],}\\
    \hspace*{5em}\textcolor{purple}{\texttt{"other\_therapy"}}\texttt{: [\{"agent": "Optune TTF"\}]}\\
    \hspace*{4em}\texttt{\}}\\
    \hspace*{3em}\texttt{\},}\\
    \hspace*{3em}\texttt{"rationale": "<brief justification>"}\\
    \hspace*{2em}\texttt{\}}\\
    \hspace*{1em}\texttt{]}\\
    \texttt{\}}
    \end{tcolorbox}
    \caption{\textbf{System Prompt Template for the Therapy Policy Agent.} The prompt is engineered to enforce clinical constraints ($\Omega$), handle temporal context mapping, and execute the feedback-driven Inverse Survival Evaluation loop.}
    \label{fig:system_prompt}
\end{figure}

\end{document}